# Joint Characterization of Spatiotemporal Data Manifolds

**Daniel Sousa[1*], Christopher Small[2]**

[1]Department of Geography, San Diego State University, San Diego, CA, USA

[2]Lamont-Doherty Earth Observatory, Columbia University, Palisades, NY, USA

**\* Correspondence:**
Daniel Sousa
dan.sousa@sdsu.edu

**Abstract**

Spatiotemporal (ST) image data are increasingly common and frequently high-dimensional (high-D). Modeling ST data can be a challenge due to the plethora of processes, both independent and interacting, which may or may not contribute to the measurements. Characterization can be considered the complement to modeling by helping guide assumptions about generative processes and their representation in the data. For high-D signals, dimensionality reduction (DR) is a frequently implemented type of characterization designed to mitigate the effects of the so-called "curse of dimensionality". For decades, Principal Component (PC) and Empirical Orthogonal Function (EOF) analysis has been used as a linear, invertible approach to dimensionality reduction and spatiotemporal analysis. Recent years have seen the additional development of a suite of nonlinear DR algorithms, frequently categorized as "manifold learning". Here, we explore the idea of joint characterization of ST data manifolds using the PC/EOF approach alongside two nonlinear DR approaches: Laplacian Eigenmaps (LE) and t-distributed stochastic neighbor embedding (t-SNE). Starting with a synthetic example and progressing to global, regional, and field scale ST datasets spanning roughly 5 orders of spatial magnitude and 2 orders of temporal magnitude, we show these three DR approaches can yield complementary information about the topology of ST data manifolds. Compared to the PC/EOF projections, the nonlinear DR approaches yield more compact manifolds with decreased ambiguity in temporal endmembers (LE) and/or in spatiotemporal clustering (t-SNE), compared to the relatively diffuse TFS produced by the PC/EOF approach. However, these properties are compensated by the greater interpretability of PCs and EOFs than of the LE or t-SNE dimensions, as well as significantly lower computational demand and diminished sensitivity to spatial aliasing for PCs/EOFs than LE or t-SNE. Taken together, we find the joint characterization using the three complementary DR approaches capable of providing substantially greater insight about the generative processes represented in ST datasets than is possible using any single approach alone.

## 1    Introduction

From agriculture to coastal erosion, and from vehicle traffic to disease transmission, many phenomena on Earth's surface are inherently *spatiotemporal* (ST): that is, variable across both space and time. But despite the ubiquity of ST processes, meaningful quantitative analysis has been limited for centuries by observational and computational capacity (Eshel, 2011; Christakos, 2017). Fortunately, in recent years drastic reductions in costs to sense, transmit, store, and process ST observations have inverted this centuries-old paradigm, leading to the dawn of the era of so-called "Big Data".

The data analysis landscape has thus fundamentally shifted: today, ST observations abound, and scientists are in need of effective and efficient tools to analyze pattern, inform modeling, and ultimately discriminate between the signal and the noise. This asymmetry between the volume of observations and capacity of inference tools is imperfectly represented by a comparison of Google Ngram word usage for 'big data' versus 'spatiotemporal' (Figure 1). It is perhaps an illustrative coincidence that the visual departure of the 'big data' usage curve occurs in 2008, the year the Landsat satellite image archive was made freely available to the public (Woodcock et al., 2008).

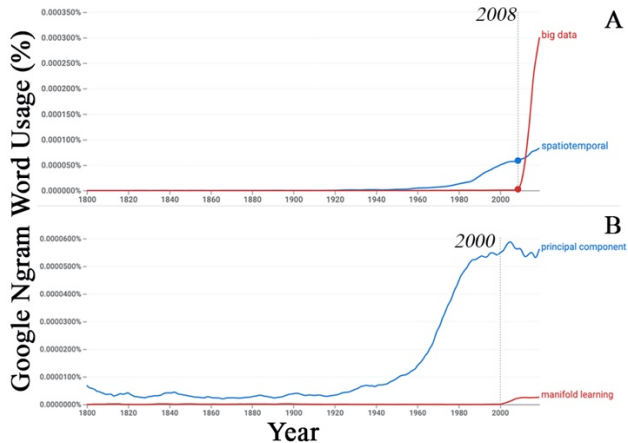

Figure 1. Historical word usage for key terms. English usage of the term 'spatiotemporal' has gradually increased for decades, while usage of 'big data' has sharply spiked since 2008 (A). Note the timing of the publication of Lorenz's scientific report on Empirical Orthogonal Functions and Weather Prediction (1956) and the opening of the Landsat archive (2008). Similarly, the term 'principal component' has steadily increased since Karl Pearson's 1901 publication, but usage of the term 'manifold learning' has been a more recent phenomenon.

ST processes frequently possess some (or all) of a suite of challenging properties. Such characteristics include high dimensionality, possibility of both abrupt and gradual changes (Verbesselt et al., 2010), spatial and temporal autocorrelation (Henebry, 1995), and cross-variable coupling (Lotsch et al., 2003). Together, these factors present formidable analytic challenges which have driven over a century of development and refinement of analytic tools (Pearson, 1901; Lorenz, 1956; Ng et al., 2002; van der Maaten and Hinton, 2008).

Characterization of ST signals complements modeling by informing presuppositions about the number, identity, and interaction of real-world generative processes which may be imperfectly represented by a set of observations (Small, 2012). Accurate characterization can be challenging for high dimensional (high-D) ST signals, due to a range of properties of high-D spaces popularly deemed the "curse of dimensionality" (Bellman, 1957). One approach to this challenging problem of ST characterization focuses on analysis of the temporal feature space (TFS). Here, low-D projections of high-D ST data are visualized and analyzed to characterize their underlying geometric and topological structure. Once characterized, this structure can then be used to design parsimonious, well-posed inverse models (Small, 2012). Conceptually, geometry implies algebra.

The TFS has been used to produce accurate characterization of complex spatiotemporal processes as diverse as oak woodland drought stress (Sousa and Davis, 2020), post-cyclone mangrove recovery (Small and Sousa, 2019), dynamics of commercial (Sousa and Small, 2019) and smallholder (Small, 2012) agriculture, cloud forest phenology and disturbance response (Sousa et al., 2019), urban development and nighttime light (Small et al., 2018), and pathogen transmission (Small and Sousa, 2021). Currently, TFS characterization relies on estimating the global variance structure of an ST dataset using linear, mutually orthogonal, variance-ordered basis functions, i.e., Principal Component (PC) and Empirical Orthogonal Function (EOF) analysis (Pearson, 1901; Lorenz, 1956). For the remainder of the text, we use the term "PC/EOF" to refer to this approach to both reflect the intrinsic complementarity between PCs and EOFs, and to be inclusive of the range of nomenclature used by various subfields throughout the literature.

While effective in many ways, the linear PC/EOF approach alone has significant limitations. First, signals that comprise real ST data are rarely truly orthogonal, so observed signals generally must be represented by linear combinations of two or more PCs/EOFs. Second, PCs/EOFs are based solely on overall (global) variance of the entire dataset and do not leverage potentially important local scale manifold topology and connectivity structure that may be present in a high-D TFS.

These limitations suggest that the information provided by PCs/EOFs might be complemented by additional dimensionality reduction tools which do not require strict linearity and/or are able to capture local feature space topology. These desired properties suggest the field of *manifold learning* as a potentially useful complement. Manifold learning refers to a class of nonlinear dimensionality reduction (DR) techniques specifically designed to focus on the connectivity structure of the feature space via observations that are deemed proximal according to a chosen statistical distance metric. The properties of this local connectivity structure are then examined, giving information which complements the global variance information given by PCs and their corresponding EOFs.

The popularity of nonlinear DR techniques has increased in recent years with improvements in computational capacity and open-source software packages like *scikit-learn* (Pedregosa et al., 2011). Such techniques are highly general, spanning a wide range of use cases. In spatiotemporal analysis, examples of nonlinear DR include Earth system modeling and data assimilation (Safaie et al., 2017), land cover classification from satellite imagery (Yan and Roy, 2015; Zhai et al., 2018), hyperspectral imagery of the water column (Gillis et al., 2005), human mobility (Watson et al., 2020), medical imaging (Kadoury, 2018), face recognition in video data (Hadid and Pietikäinen, 2009), and more.

This analysis explores the approach of joint characterization introduced by (Sousa and Small, 2021) for the case of spatiotemporal analysis of image time series. The approach used here focuses on the complementarity of linear and nonlinear dimensionality reduction algorithms. Each algorithm emphasizes a different aspect, or variance scale, of signal within a ST dataset. Here, we implement one linear method (PCs/EOFs) and two nonlinear methods (Laplacian Eigenmaps, LE; and t-distributed stochastic neighbor embedding; t-SNE). Because t-SNE has a stochastic element that makes a single realization effectively non-repeatable, we use use a Monte Carlo approach in which the low order PCs of a number of independent 2D t-SNE realizations, PC(t-SNE), are used to characterize consistently recurring structure within the t-SNE feature spaces (Small and Sousa, 2021). While we focus on these two nonlinear methods due to their popularity and complementarity, we note that DR is an active area of research – new approaches continue to be developed, and other existing approaches may be more appropriate for a given application. For these reasons, the purpose of this work is not to provide a specific methodological recipe, but rather a general conceptual framework with which to consider analysis of ST processes. Beginning with a synthetic example and progressing to global, regional, and field scale observational datasets, we ask the following questions:

1) Which geometric and topological characteristics does each dimensionality reduction algorithm accentuate or suppress when producing a low-D representation of the topological structure of a high-D ST dataset?

2) How do the properties of each algorithm map onto strengths and weaknesses for specific ST analysis applications?

3) In what ways can methods be used together to yield a more effective characterization than is possible with any single method alone?

In addressing these questions, we demonstrate a generalized analytic framework for joint characterization of ST data manifolds. We extend the concept of the temporal feature space beyond the linear domain, illustrating the potential for meaningful signal to exist across multiple variance scales within the same ST dataset, and for the ability of complementary DR approaches to jointly capture that signal.

## 2  Toy Example

### 2.1  Toy Example – Setup

We motivate the concept using a highly simplified synthetic dataset (Figure 2). Here, a 100 x 100 x 100 spatiotemporal data cube is created using random linear combinations of three signals: one sinusoid and two decaying exponentials with different amplitudes and decay constants.

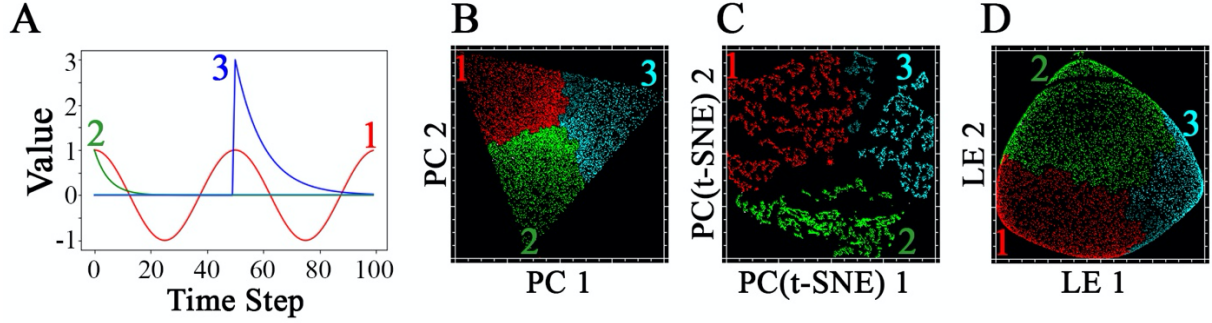

Figure 2. Illustration with synthetic data. Linear combinations of sinusoidal (A; 1) and decaying exponential (A; 2 and 3) signals are used to generate a 100 x 100 x 100 spatiotemporal cube. Weights for each signal are randomly chosen from a uniform distribution and forced to sum to unity. PCA-based linear dimensionality reduction (B) reveals the low order temporal feature space to be occupied by a trigonal planar mixing manifold, with >99% of variance allocated to the first two dimensions. PC(tsne) yields a higher dimensionality (88% of variance in first two dimensions) and more clustered topology. Red, green, and cyan clusters identified from the PC(tsne) space effectively identify the signal with dominant weighting in each pixel - at the expense of a clear representation of the mixing continuum (C). Three clusters comprised of mixtures of signals 1 and 3 form an exception (dark cyan), mapping closer to the signal 1 group in PC(tsne) space despite slighly higher signal 3 weights. Cluster colors are applied to the other two spaces for visual comparison. In contrast, the low-order LE dimensions (D) yield a more continuous topology, with sharply defined edges and rounder but denser apexes than are present in the PC space.

The generating signals are given by the following functions (Figure 2, A):

$$s_1 = \cos\left(\frac{\pi}{25} t\right)$$

$$s_2 = e^{\frac{-t}{5}}$$

$$s_3 = \begin{cases} 0 & for\ x = [0,49] \\ 3e^{\frac{-(t-50)}{10}} & for\ x = [50,100] \end{cases}$$

Within the spatiotemporal data cube of the toy example $T$, each synthetic "observation" $T_{x,y}(t)$ is computed as a linear combination of the above three signals, with weights chosen from a uniform random distribution and forced to sum to unity.

The forward model can be expressed as a weighted linear combination as: That is,

$$T_{x,y}(t) = w_1 s_1 + w_2 s_2 + w_3 s_3, \quad such\ that: \quad w_1 + w_2 + w_3 = 1$$

The inverse problem involves the identification of the temporal endmembers ($s_i$) and the estimation of their corresponding weights ($w_i$) for each observed $T_{x,y}(t)$.

Because the spatiotemporal cube has 100 time steps, each $T_{(x,y)}$ can be considered a vector residing in 100-D space. However, the simplicity of the generating functions suggests that the true dimensionality of the underlying signals is far lower than 100-D. Geometrically, the data lie on a low-D manifold within the full high-D space. Topologically, strictly imposed linear mixing dictates that this triangular manifold is fully interconnected with sharp corners and straight edges. Such properties of the low-D manifold imply fundamental constraints on the design of a well-posed inverse problem: geometric and topologic structure imply algebraic structure. In this case, the inverse problem amounts to estimation of the relative contribution of each independent signal (temporal endmember) to a general observation potentially containing contributions from one, two or three of the signals.

We next use this synthetic dataset to illustrate the effect of complementary linear and nonlinear dimensionality reduction techniques.

## 2.2 Toy Example – Linear Dimensionality Reduction

One way of visualizing the low-D data manifold is through the long-established linear technique of PC/EOF analysis. Through this lens, the data are decomposed onto a set of mutually orthogonal eigenvectors which are sorted on the basis of variance. The power of this approach is illustrated in Figure 2B. Here, two dimensions represent $T$, together comprising >99% of overall variance in the data. The algebraic structure imposed by the linear mixing equations above maps cleanly onto trigonal planar geometric structure reminiscent of a ternary diagram. Pixels with (nearly) pure contributions from each one of the three temporal endmember (tEM) signals occupy one of the three corners of this triangle. Pixels occupying the sharp linear edges represent binary mixtures of two EMs; and more generally, pixels in the body of the triangle have a contribution from each of the three EMs linearly weighted by Euclidean distance to each corner. The variance partition and topology of the temporal feature space immediately characterize the 2D manifold and three temporal endmembers corresponding to the three input signals.

## 2.3 Toy Example – Nonlinear Dimensionality Reduction

Manifold learning techniques cast the question of characterization in a fundamentally different light (Van Der Maaten et al., 2009). Here, statistical similarity is evaluated for pairs (or more generally n-tuples) of observations, with 'similarity' defined according to the analyst's distance metric of choice. One simple metric is Euclidean distance, which can be visualized for our toy example $T$ as simple spatial proximity in the trigonal mixing space of Figure 2B. In the case of a higher-D dataset, Euclidean distance cannot be visualized accurately solely on the basis of a single 2-D scatterplot, and instead takes the form of a higher-D generalization of the underlying principle. While a wide range of analytic approaches can be used to characterize the pairwise statistical similarity structure of a dataset, here we focus on two nonlinear manifold learning algorithms: t-SNE and LE. These two algorithms leverage neighborhood connectivity information in fundamentally different ways.

LE uses a graph theoretic approach, considering the high-D observations as an interconnected network of nodes and edges. Connectivity among proximal data points can be considered using a distance threshold or number of nearest neighbors. Once the graph of the observations is constructed, its matrix Laplacian is computed and decomposed into eigenvectors. The contribution of each Laplacian eigenvector to each data point is then known and can be used to characterize the data. For more information about LE, see (Ng et al., 2002; Shi and Malik, 2000; Von Luxburg, 2007).

In the case of our toy dataset *T*, we can see that the connectivity structure of LE (Figure 2C) accurately reflects the generative spatial mixing process. Both the connectivity structure and bounding endmembers from Panel B are clearly preserved in LE (Panel D). However, the linear edges and sharp corners are rounded, a characteristic of LE which can be conceptually considered to reflect the (roughly) analogous nature of the Laplacian to geometric curvature.

In contrast, the fundamental idea underlying t-SNE is to minimize the difference between two probability distributions: one constructed over the given high-D observations and another constructed over the desired low-D map. The metric used to evaluate differences between distributions is the Kullback-Leibler divergence. The t-SNE algorithm introduced by (van der Maaten and Hinton, 2008) is a variant of the more general stochastic neighbor embedding (*SNE*) algorithm of (Hinton and Roweis, 2002). For an excellent practical introduction to t- SNE, see (Wattenberg et al., 2016).

Elements of the t-SNE output are stochastic. Specifically, the absolute location of any given observational cluster in the low-D map produced by t-SNE is essentially random, and potentially important differences in boundaries among clusters can exist among different realizations of the algorithm – as implied by Figure 2. One commonly used approach to resolve this limitation is to run multiple realizations of the algorithm and choose the result with the minimum divergence (van der Maaten, 2021). Here, following (Small & Sousa, 2021), we take a different tack. Rather than choosing the single run with minimum divergence, we stack the output of multiple t-SNE realizations and compute the PCs of this stack, to which we refer as PC(t-SNE). This has the effect of capturing cluster consistency, as observations which routinely cluster together in t-SNE space plot together in PC(t-SNE) space. The eigenvalue distribution of the PC(t-SNE) result also provides information about the global dimensionality of the data manifold(s) resolved by t-SNE.

Applied to our toy dataset *T*, we can see that PC(t-SNE) gives a fundamentally different, more clustered output (Figure 2D) than PCs/EOFs or LE. On the one hand, this clustering could be considered a strength: the relatively well-separated red, green, and blue clusters accurately identify the dominant input signal contributing to each observation. On the other hand, the same property could be considered a weakness, introducing granularity to the low-D map which is not necessarily representative of any underlying generative process in the spatiotemporal dataset. One question we seek to address here is the consistency with which t-SNE clusters resolve known differences among subsets of observations. While the separability of the t-SNE clusters shown in Figure 2 generally corresponds to the largest endmember fraction in the mixture, we highlight the exception (in dark cyan).

### 3. Global Scale Example – Climate Reanalysis

We now perform a joint characterization of a commonly used set of ST observations: gridded climate reanalysis data produced by the Climatic Research Unit (CRU) at the University of East Anglia (UEA) (Mitchell and Jones, 2005). Here, we apply these DR algorithms to 20 years (1983-2002) of monthly estimates of mean temperature and total precipitation (T+P).  Each pixel time series represents 102 x 12 monthly mean temperatures and total monthly precipitation estimated for each 1°x1° grid cell on land, excluding Antarctica. The resulting maps (Figure 3a) of the low-D spaces illustrate complementarities among the DR approaches. Traditional PC/EOF DR (top) captures broad zonal and longitudinal climatic patterns, representing the vast majority of points as occupying a continuum spanning differing endmember climate zones. In contrast, t-SNE (lower right) represents the global terrestrial T+P TFS as an amalgamation of highly discretized subregions, conceptually resembling a choropleth map. Some features, like the longitudinal Eurasian dipole, are accentuated in t-SNE that are either relegated to higher dimensions or represented as a linear combination of two or more PC/EOF dimensions. LE (lower left) complements both of these two approaches, representing the global T+P space as fundamentally comprised of continuous nonlinear gradients which are notably sharper than present in the PC/EOF space, but lack the discrete clustering of the t-SNE space. Discrete physiographic features like the Rocky, Himalayan, and Andean mountain ranges emerge clearly using PCs/EOFs and LE, but not in t-SNE. Comparison of PC/EOF, LE and t-SNE also reveals that even the low order dimensions of PCs/EOFs resolve finer scale, but physically meaningful, distinctions such as the orographic high precipitation regions of the Western Ghat in India and the Chittagong Hill Tracts in Bangladesh. These features are resolved as distinct clusters in individual t-SNE realizations, but not in the low order dimensions of PC(t-SNE). They appear within the continuum of the LE TFS.

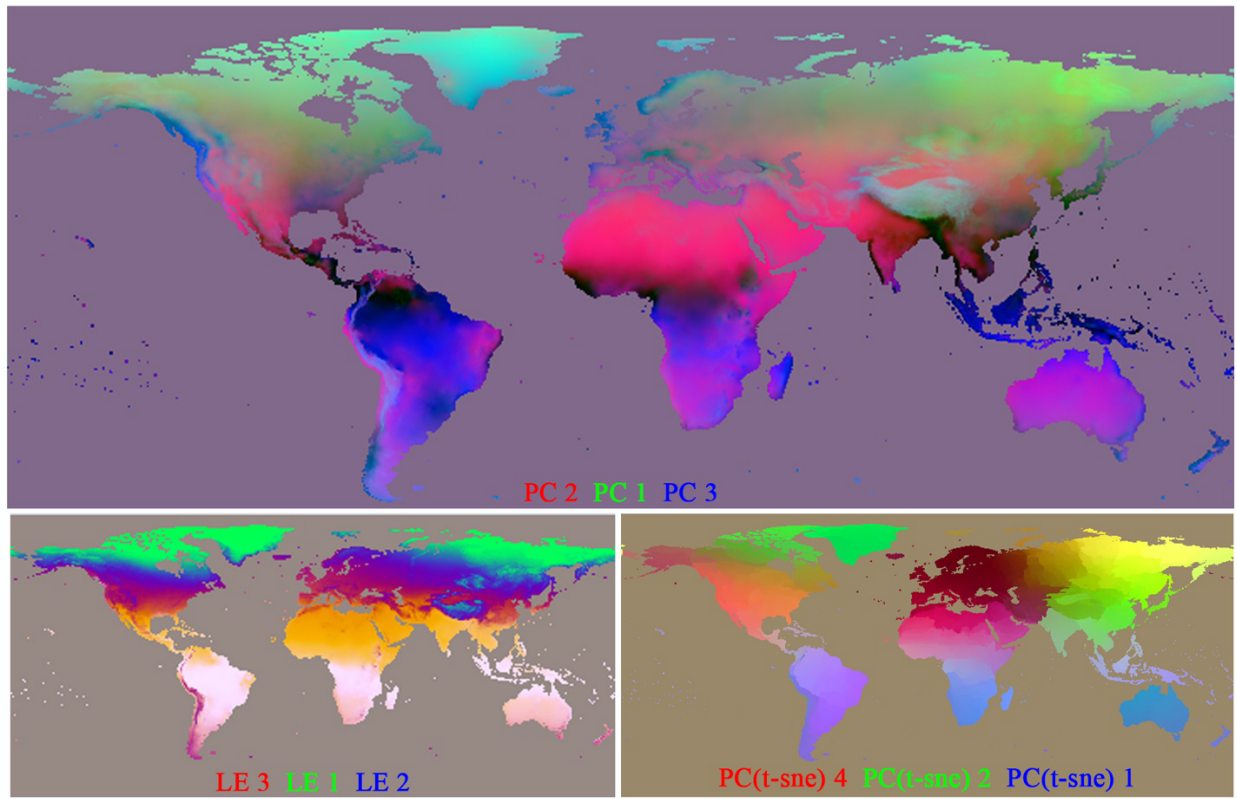

Figure 3a. Joint characterization yields complementary spatiotemporal patterns when applied to a century of global climate reanalysis data. Characterization of monthly mean air temperature and total precipitation estimates for 100 years (1900-2000) using linear and nonlinear dimensionality reduction reveals both continuous gradation and discrete clustering. Linear variance-based decomposition (PCA, top) captures broad bioclimatic patterns. Non-linear approaches produce complementary maps: Laplacian Eigenmaps (bottom left) accentuates gradients, with clear separation among tropical, arid, temperate, and boreal climates in the three lowest dimensions. In contrast, a compilation of 100 t-SNE runs captures both connectivity relationships and clearly identifiable geospatial clusters.

Temporal feature spaces derived from each approach (Figure 3b) illustrate key differences in geometry and topology. PCs/EOFs (top) represents the majority of grid cells as occupying a position within a continuum bounded by endmember climates. The PC1 vs PC2 feature space effectively discriminates between hot & dry, hot & wet, and cold climatic endmembers, bearing a strong resemblance to the mean temperature vs precipitation space on which terrestrial biomes are defined, e.g. in (Small and Sousa, 2016). Density shading reveals prevalence of linear and nonlinear mixing relationships, as well as sparse data representation of high variance patterns with low PC 2 and low PC 3 values. Points with maximum PC 1 values differentiate along the PC 3 axis between Polar Ice Cap (Köppen Climate Type EF, Greenland) and Polar – Tundra (Köppen Climate Type EF, Siberia and northern Canada).

In contrast, the clustering evident in the t-SNE map is clearly represented in the corresponding low-order TFS (bottom row). Here, the ST data manifold is represented as a discretized nonlinear continuum, with point density spread in a relatively uniform manner among clusters. Mixing relationships are similar to those found with PCs/EOFs, but effectively collapsed onto a narrow manifold rather than diffusely spread throughout the low-order TFS. It is noteworthy that the small isolated clusters in the t-SNE space correspond to geographically isolated locations (e.g. Madagascar, Australia, Greenland) as well as topographically isolated regions (e.g. Altiplano, Anatolia, Tibetan Plateau), providing a clear physical basis for distinction from the more continuous climatic gradients. The LE TFS (center row) possesses a fundamentally distinct geometric structure. Here, the global terrestrial T+P space is represented as a tight manifold with a nonlinear continuum spanning a low dimensional surface bounded by three unambiguous endmembers corresponding to those bounding the PC1/PC2 feature space. For this popular ST dataset, each of the three DR approaches clearly provides information that the other two do not, illustrating the potential complementarity benefits of joint characterization of the global climate space.

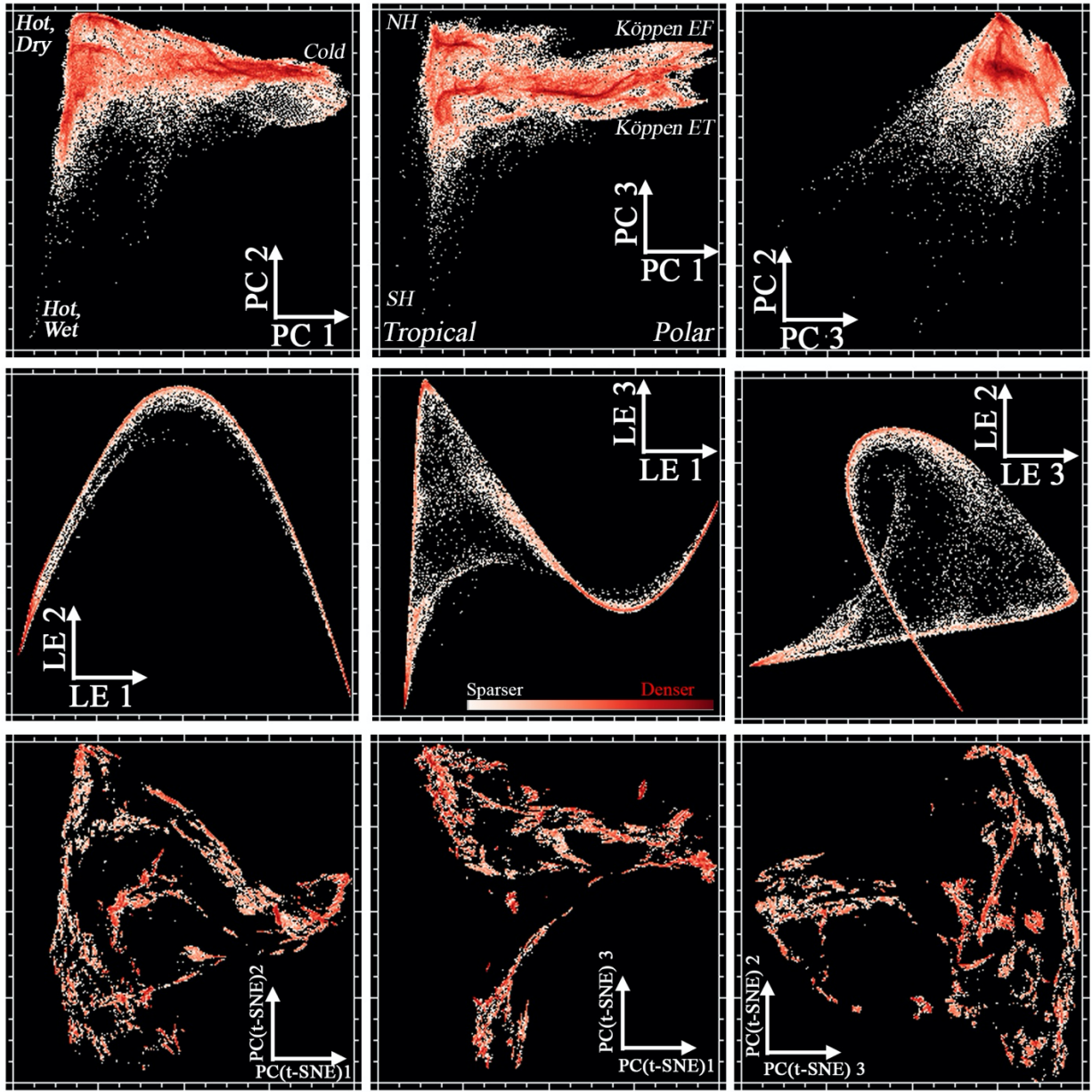

Figure 3b. Joint characterization yields complementary temporal feature spaces when applied to a century of global climate reanalysis data. The continuous gradation and discrete clustering observed in geographic space in Figure Xa maps onto temporal feature space structure. Linear variance-based decomposition (PCA, top row) captures dominant patterns of overall variance across the entire domain, yielding a diffuse point cloud with substructure indicative of regional-scale gradients. Non-linear approaches produce complementary results: Laplacian Eigenmaps (center row) captures a continuous helicoid manifold that clearly expresses endmember climates and mixing relations, PC(t-SNE) captures both connectivity relationships and clearly identifiable clusters. Each temporal feature space emphasizes different aspects of the geometric and topological structure of the spatiotemporal data manifold.

### Regional Scale Example – Sahel MODIS

Next, we investigate the ability of joint characterization to provide useful information across spatiotemporal scales and generative processes. Specifically, we explore rainfall-driven vegetation phenology across the African Sahel. This spatiotemporal cube is a time series of MODIS Enhanced Vegetation Index (EVI) vegetation abundance maps in which each pixel time series represents the aggregate vegetation phenology within a 250x250 m footprint. Relative to the global climate illustration above, these data are ~2x more frequent (16-day composite) and ~400x spatially finer (250 m).

The TFS resulting from each DR approach is shown in Figure 4a. PCs/EOFs identify clear temporal endmembers corresponding to barren (EM1) and evergreen (EM2) phenologies, connected by a diffuse helical manifold of seasonal grass/shrub vegetation driven by latitudinal migration of the InterTropical Convergence Zone (ITCZ) and associated rainfall. EOF 1 controls the amplitude of the seasonal vegetation abundance.  EOFs 2 and 3 reveal complementary out-of-phase sinusoids which together combine to form a phase plane. An additional double monsoon signal is seen in EM3, distinct from the single annual signal.

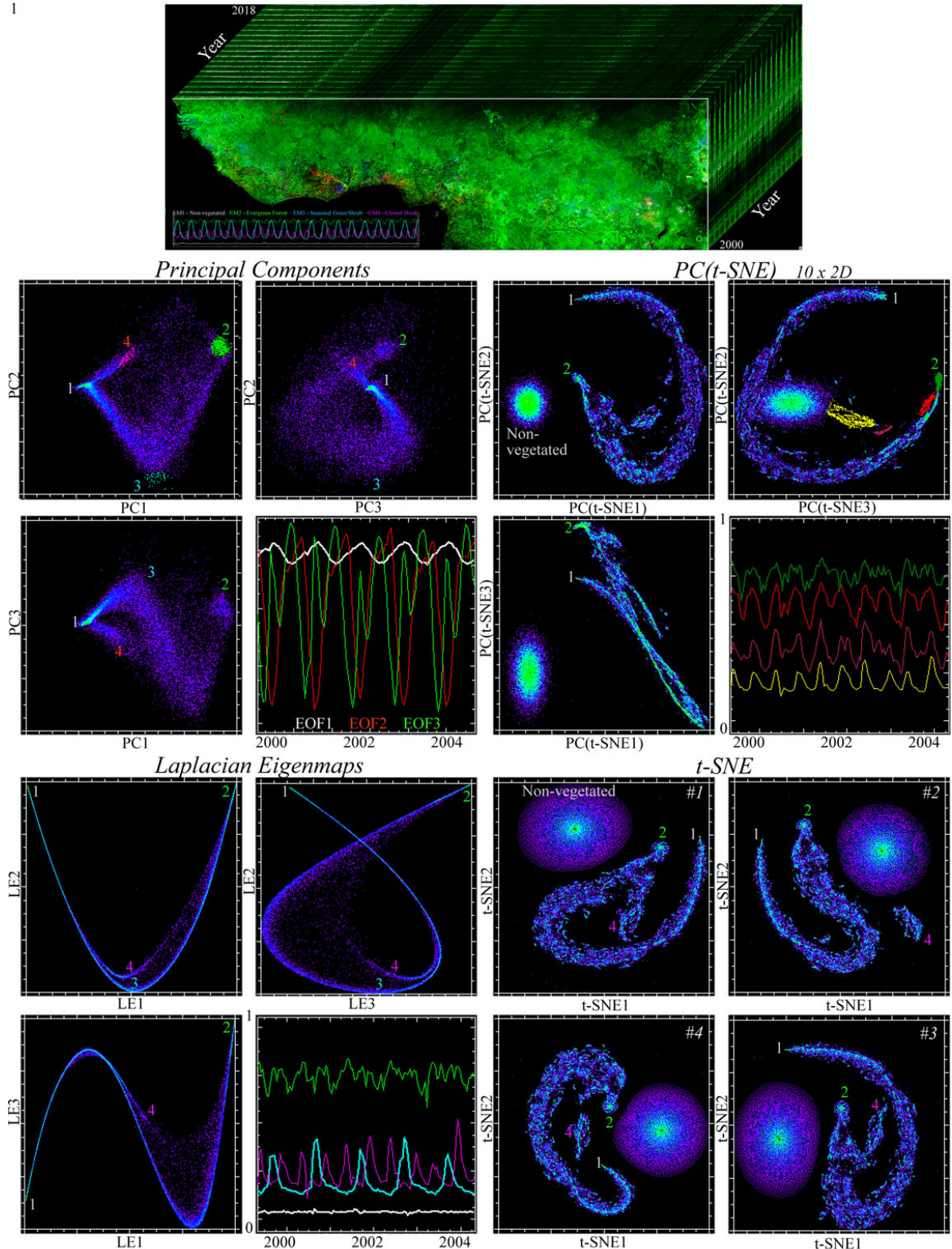

Figure 4a Complementary temporal feature spaces for Sahel MODIS EVI vegetation phenology. Principal components yield a diffuse continuous helical feature space in which PC1 controls amplitude while PC2 and PC3 form a phase plane to represent the seasonal shift in greening and senescence following the latitudinal shift of precipitation. Temporal endmembers (inset) from apexes of the PC cloud form a convex hull bounding the full range of phenologies. Temporal EOFs show the latitudinal phase shift and isolated double monsoon seen in EM3 and EM4. Laplacian Eigenmaps yield a much less diffuse, more continuous feature space with similar structure and EMs. Individual t-SNE spaces vary in shape and orientation but with similar topology, also showing an amplitude continuum with distinct clusters for non-vegetated areas and closed shrublands fed by the Somali double monsoon (EM4). PC(t-SNE) of 10 2D t-SNE realizations produces a more compressed helical structure, also varying continuously in amplitude and phase. In comparison to the individual t-SNE realizations, the PC(t-SNE) topology shows less divergence with increasing amplitude along the continuum with more distinct clustering of the double monsoon phenologies.

Nonlinear dimensionality reduction methods identify a much more compact low-D manifold, emphasizing some parts of the structure of the linear space in each case. LE collapses the EM1 – EM3 – EM2 continuum into a single, well-defined curvilinear helicoid, with double monsoon EM3 diverging from the primary manifold near its midpoint. PC(t-SNE) identifies a similar overall manifold structure, but with much more defined clustering. Unvegetated pixels form a large, diffuse cloud in PC(t-SNE) space. PC(t-SNE) also separates coherent clusters which are not clearly defined using either PC/EOF or LE methods alone. The stochastic nature of individual t-SNE realizations is illustrated through the example realizations shown in the bottom right quadrant of the figure.

The geometry and topology from Figure 4a can then be leveraged to design a temporal mixture model of the vegetation phenology. Here, we illustrate a simple linear model constructed using four temporal endmember phenologies identified from the TFS characterization (Figure 4b inset). The resulting phenology map, shown in Figure 4b, is effective at capturing a continuum of distinct phenological zones, as well as variability within and among those zones.

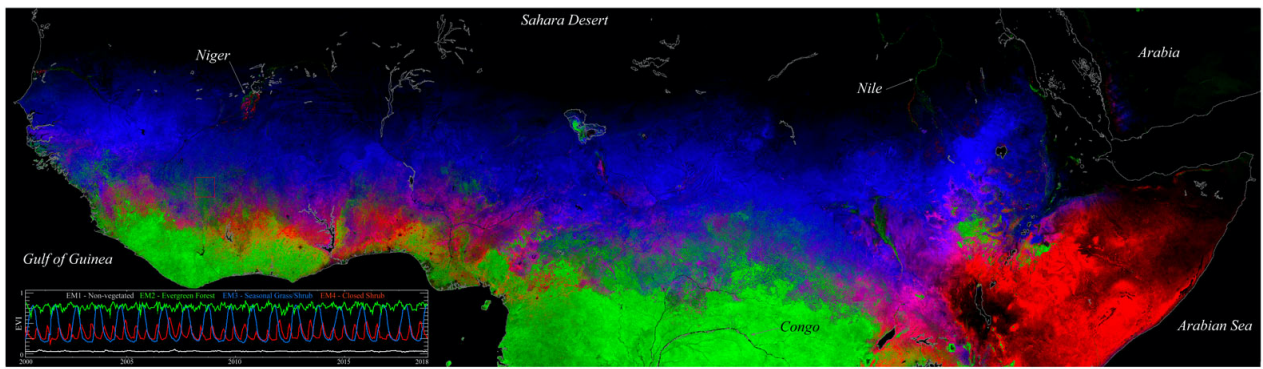

Figure 4b Sahel vegetation phenology map derived from inversion of a temporal mixture model of MODIS EVI time series . Linear combinations of the temporal endmembers (inset) derived from the temporal feature space in Fig. 4a represent vegetation communities in each 250 m pixel time series as mixtures of evergreen trees, seasonal grasses and shrubs. The latitudinal gradient from evergreen forest to seasonal grasses and shrubs is a result of seasonal variation in precipitation following the north-south oscillation of the InterTropical Convergence Zone. The Niger inland delta and Nile River Valley phenologies contrast their surroundings because the seasonality of catchment discharge is delayed relative to downstream precipitation timing. The 4 endmember temporal mixture model represents 95% of the MODIS EVI time series with < 10% misfit.

### Field Scale Example – California Agriculture

We next extend the investigation of scale dependence to a study area relevant to practical land management: field-level agricultural mapping in California (Figure 5). This extends our investigation of scale dependence by covering a ~20x shorter record (1 year) with ~3x more frequent revisit (3-5 day) and 1000x finer spatial scale (9 km$^2$) than was present for the MODIS example above. The study area here is an orchard-dominated 3 km x 3 km subset of Kern County, the highest-value crop producing county within California's agriculturally diverse Central Valley (Kern County, 2019). Here, we use subpixel vegetation fraction estimated from 30 m Harmonized Landsat-Sentinel (HLS) multispectral imagery by inversion of a linear spectral mixture model using generalized global spectral endmembers from (Small, 2018).

The area shown here is dominated by two high-value orchard crops: almonds and pistachios. A publicly available county-level crop map (Kern County, 2019) provides ground truth of crop type (far left, top; superimposed on false color reflectance image). Average vegetation fraction time series of each crop (far left, bottom) show the almond crop is characterized by earlier, higher amplitude green-up and more rapid senescence than the pistachio crop. Two pistachio fields (white in reflectance image) behave differently, with lower overall vegetation fraction presumably due to age of orchard.

PCs/EOFs (center left) generally capture these differences, with almonds and pistachios forming diffuse clusters largely separable on the basis of PC/EOF 2. The two distinct pistachio fields plot as a smaller cluster closer to the center of the linear feature space. However, the diffuse topology of the data in linear feature space yields ambiguity in the optimal location of decision boundaries. In effect, within-class variance is accentuated at the expense of between-class variance.

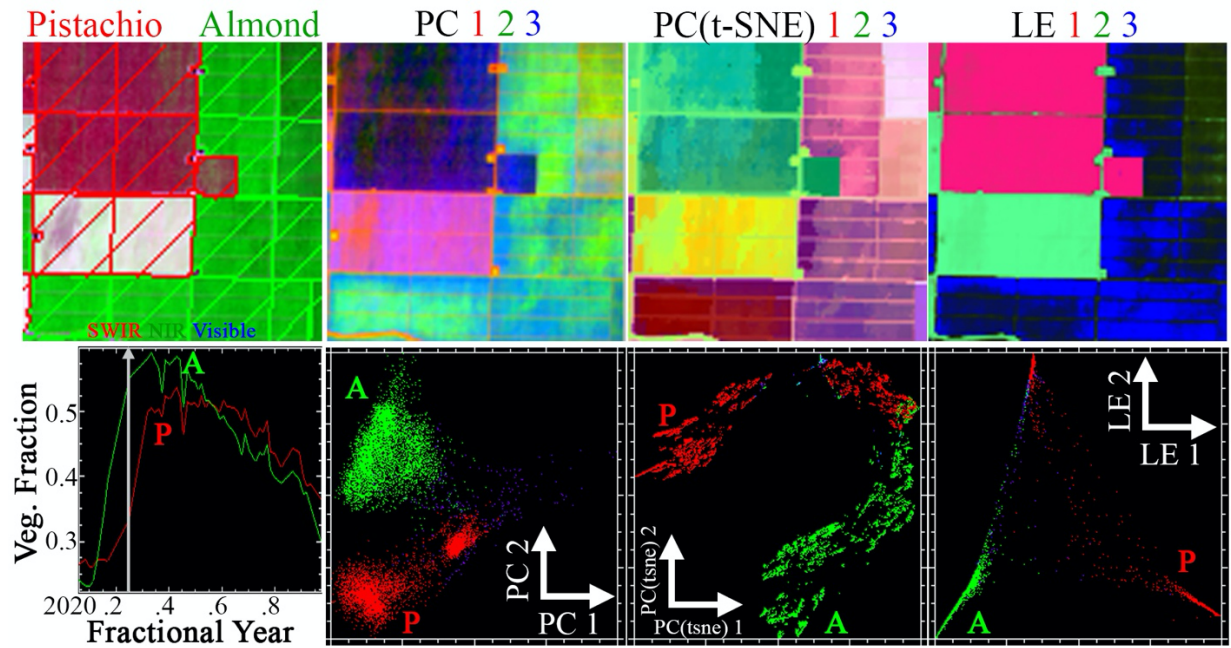

**Figure 5. Dimensionality reduction techinques capture complementary aspects of crop phenology.** Other than roads and structures, a 3 km x 3 km subset of Kern County, CA is entirely comprised of almond and pistachio orchards (upper left; image date Mar 23, 2020). Mean time series of all pixels from each crop (lower left) reveals similar phenology, but with earlier and greener leaf-on maximum, and more rapid senescence, for almonds (A) than for pistachios (P). Each technique resolves spatially coherent (top) patterns which resolve variance both within and among classes. The locus of convergence in PC(tnse) space corresponds to the top apex in LE space, which features a lower amplitude phenology indicative of a younger orchard with smaller, less full canopies.

Nonlinear dimensionality reduction algorithms capture different aspects of the high-D data structure. Almond and pistachio fields are sharply distinct as well-separated (Transformed Divergence = 2.0) apexes in LE space, with the two distinct pistachio fields forming a third well-separated apex at high LE 2. In contrast, PC(t-SNE) captures a continuously varying nonlinear data manifold. Almonds and pistachios form opposite ends of this manifold, with the two distinct fields plotting near the lengthwise center of manifold.

## Discussion

### 6.1 Strengths and Limitations

Any low-dimensional representation of a high-dimensional space will generally result in retention of some subset, and loss of some other subset, of the total high-D information. The analysis presented here illustrates complementarity among three philosophically differing approaches to achieving such low-D representations, focusing on the special case of spatiotemporal characterization. Depending on the application, the subset of information retained by each approach may be more or less useful. In many cases, comparative analysis facilitated by joint characterization using multiple DR approaches may prove to be more useful than the sum of its parts.

Specifically, important differences among algorithms manifest as differential sensitivities to scales, geometries, and topologies of variance. Perhaps the broadest difference is between the global and local scales of operation between the linear and nonlinear DR approaches. The linear (PC/EOF) approach implemented here operates on the overall global variance structure of the ST dataset, allowing each observation explicit contextual representation in terms of the global space, but sacrificing potentially important local structure for generality. In contrast, both of the nonlinear approaches (LE and t-SNE) focus on statistical similarity structure among observations at the expense of downweighing global structure. This difference in formulation of the DR problem maps directly onto geometry and topology of each low-order Temporal Feature Space. The linear, global TFS yields relatively diffuse point clouds due to its strict orthogonality requirement and usage of a metric of global dispersion (variance). In contrast, the low-order TFS for each nonlinear algorithm yields a visibly tighter manifold, with decreased ambiguity in temporal endmembers (Laplacian Eigenmaps) and/or in spatiotemporal clustering (t-SNE) – a direct result of optimizing metrics focused on local structure at the expense of global structure.

However, neither algorithmic strengths nor limitations occur in isolation. For instance, the diffuse nature of the PC/EOF TFS poses a challenge, but is compensated by the explainable connection between spatial and temporal dimensions provided by explicit EOF time series associated with each PC spatial pattern. This relative ease of interpretability is not a feature of LE or t-SNE. Similarly, the unambiguous clustering relations identified by t-SNE are in some sense a feature, for instance allowing clear determination of decision boundaries for discrete classification; but can also be a limitation by artificially introducing clustering that may not be present in the generative processes, as observed in the synthetic example of Figure 2. As the degree of clustering can be influenced by the Perplexity parameter chosen, in some cases a sensitivity analysis may be warranted to quantify the persistence of specific clusters. Similarly, LE is effective at removing ambiguity in endmember ST patterns, but is limited in explainability. Further, because both t-SNE and LE operate on similarity relations among statistically similar observations, both are considerably more sensitive to spatial aliasing than PCs/EOFs, which treat each observation as independent of all others. Notably, this is not necessarily true for temporal aliasing, as the symmetry between spatial and temporal dimensions in broken in LE and t-SNE.

### 6.2 Broad Efficacy

The global, regional, and field scale examples shown above span roughly 5 orders of magnitude spatially and 2 orders of magnitude temporally, illustrating the ability of joint characterization to provide useful information at global, local, and regional scales. In particular, the approach is shown to be effective in both characterizing processes occurring at scales coarser than pixel resolution (field-level agriculture, broad climate gradients) and their associated subpixel spatial mixing processes (smallholder Sahel agriculture). Further, joint characterization is shown to yield useful results for disparate biogeophysical ST processes including climate as well as natural and human-managed vegetation. This broad efficacy suggests considerable potential for relevance to other non-ST but high-D data types, for instance to the field of imaging spectroscopy as recently suggested in (Sousa and Small, 2021).

**6.3    Synthesis**

The concept of the temporal feature space was initially introduced as the basis for characterization of high dimensional ST data to inform the feasibility and design of temporal mixture models.  By providing a model-agnostic projection of the temporal feature space, the combination of eigenvalue variance partition, temporal EOFs and spatial PCs, the Singular Value Decomposition of ST data informs the 1) spatiotemporal dimensionality, 2) linearity (or lack thereof) and 3) temporal endmember selection (Small, 2012).  While individual temporal EOFs can be informative, it is their linear combinations that represent the actual data.  Individual EOFs almost never act alone.  As such, the temporal endmembers identified from the TFS that provide a more physically meaningful set of basis functions on which to project the higher dimensional ST data.  If a TFS structure can be accurately represented (as quantified by model misfit to observation and positivity of EM fractions) with a relatively small number of temporal EMs, a temporal mixture model can provide a useful abstraction of high dimensional ST data.  In addition, the variance partition given by the eigenvalues of the covariability matrix can provide a basis for distinguishing between deterministic and stochastic components of the ST data (Lorenz, 1956; Small, 2012).  The only assumptions implicit in the use of Principal Components and EOFs are that variance corresponds to information content and that correlation corresponds to redundancy.  While these are often valid assumptions, they also bias the projection of the temporal feature space accordingly.  The potential value of supplementing the global variance structure given by the PCs/EOFs with nonlinear manifold learning tools is the potential for preservation of low variance information content that may be lost in the higher order dimensions of the PCs/EOFs.  In other words, the assumptions implicit in some nonlinear manifold learning approaches are complementary to the assumptions on which PC/EOF analysis is based.  As shown in the examples presented here, LE and t-SNE are complementary in that the LE spaces are simpler, less diffuse and more continuous than the PC-based TFS, while the t-SNE (and PC(t-SNE)) spaces can be both continuous and clustered with verifiably meaningful clusters not resolved by the PC-based TFS (Sousa and Small, 2021; Small and Sousa, 2021).

**6.4    Practical considerations**

Even for highly effective approaches, practical limitations can severely restrict use by other analysts. Fortunately, code facilitating each of the DR algorithms used in this study is freely available from numerous open source web repositories, most notably Python's scikit-learn. Despite code availability, however, computational limitations remain a key chokepoint for easy adoption, especially for the nonlinear algorithms. Both mathematical elegance and decades of computer science have facilitated algorithms enabling efficient PC/EOF analysis of datasets with millions of spatial observations, each with hundreds of time steps, on typical personal computing hardware. The same is unfortunately not (yet) true of t-SNE and LE. Because of the nature of such manifold-based approaches, computational resources scale roughly with the square of the number of spatial – but not temporal – samples. The key limitation is generally sufficient RAM to store large matrices of pairwise (or n tuple-wise) metrics. For reference, a Lenovo laptop running x64 Linux with 32 GB of RAM and scikit-learn version 0.24.2 was able to compute both LE on datasets with a maximum of roughly 25,000 spatial samples, and t-SNE on datasets with a maximum of roughly 250,000 spatial samples. Temporal depth was not found to be a limitation for either algorithm for the datasets examined here.

Notably, other linear and non-linear DR approaches are also worthy of consideration: e.g., ISOMAP, Locally Linear Embedding, Hessian Eigenmapping, Local Tangent Space Alignment, metic and non-metric multidimensional scaling, UMAP, and more. Because the purpose of this work is to introduce the idea of the joint characterization for spatiotemporal data manifolds, rather than present a thorough intercomparison of all available methods, we defer detailed examination of these other approaches to future work. However, we do note that in some sense PCs/EOFs, Laplacian Eigenmaps, and t-SNE can be argued as representing endmember DR philosophies and thus covering a substantial subset of the current DR landscape.

For some applications, algorithm automation is a primary concern. We note here that in a fundamental sense, the joint characterization approach is inherently non-automated. This is because joint characterization requires the scientist to utilize visual perception and critical thinking in exploratory data analysis and interpretation of geometric and topological TFS structure. That being said, the information provided by joint characterization can then be used to design parsimonious inverse models which could potentially be fully automated.

### Conclusions

This analysis explores the utility of joint characterization of spatiotemporal data using synthetic, reanalysis, and satellite datasets. Joint characterization is implemented using a set of three linear and nonlinear approaches to dimensionality reduction: Empirical Orthogonal Functions, Laplacian Eigenmaps, and t-distributed Stochastic Neighbor Embedding. The three approaches are shown to yield complementary characterizations for a variety of generative processes at global, regional, and field scales. The results of this analysis suggest that joint characterization can provide a useful framework to visualize the geometric and topologic structure of high dimensional spatiotemporal data manifolds to assist with the design of effective and parsimonious inverse models, as well as improved class separability for discrete thematic classifications.

### Acknowledgements

DS thanks Jeffrey Bowles, Phil Broderick, Kerry Cawse-Nicholson, Frank Davis, Chip Miller, Dave Schimel, David R. Thompson, Phil Townsend, and Ting Zheng for fruitful and interesting conversations.

### Conflict of Interest

The authors declare that the research was conducted in the absence of any commercial or financial relationships that could be construed as a potential conflict of interest.

### Author Contributions

Study conception: DS and CS. Computation: DS. Analysis & Interpretation: DS and CS. Draft manuscript preparation: DS. Manuscript revision: DS and CS. All authors reviewed the results and approved the final version of the manuscript.

### Funding

Work done by DS was funded by the taxpayers of the State of California.  CS was funded by the endowment of the Lamont Doherty Earth Observatory.